# An AGI with Time-Inconsistent Preferences

May 28, 2019

James D. Miller and Roman Yampolskiy

Jdmiller@Smith.edu, roman.yampolskiy@louisville.edu

## Introduction

This paper reveals a trap for artificial general intelligence (AGI) theorists who use economists' standard method of discounting. This trap is implicitly and falsely assuming that a rational AGI would have time-consistent preferences. An agent with time-inconsistent preferences knows that its future self will disagree with its current self concerning intertemporal decision making. Such an agent cannot automatically trust its future self to carry out plans that its current self considers optimal.

Economists have long used utility functions to model how rational agents behave (see Mas-Colell et al., 1995). AGI theorists often rely on these utility functions because they assume that an AGI would either start out as rational or modify itself to become rational (see Omohundro, 2008; Yudkowsky, 2011; Bostrom, 2014; Soares, 2015; Yampolskiy, 2016).

When economists model intertemporal decision making, they assume that people place a lower value on receiving money or utility in the future than they do today because people discount future rewards. Economists generally assume that such discounting takes on a particular functional form. Critical for this paper, this functional form causes agents to have time-consistent preferences, and this form does not follow from the assumptions of rationality.

This paper explains what time-consistent preferences are, discusses why rational AGIs will likely not have them, and explores how an AGI with time-inconsistent preferences might behave. Finally, the paper considers how time dilation under Einstein's theory of relativity could complicate an AGI's efforts to fix its time-inconsistency problem.

## Discounting

The standard discounting function economists use assumes that discounting takes the form of $\delta^t$ where δ is an exogenously determined parameter between zero and one, and t is how many periods in the future the agent expects to receive the money or utility (see Frederick et al., 2002: 358; the exact notation used can differ from what is presented in this paper). The lower the value of δ, the more the agent discounts a reward expected to be received in the future. The present value to an agent of knowing that it will receive, say, $9 in period t is $9$\delta^t$. An agent is indifferent between receiving $9$\delta^t$ immediately or receiving an absolute guarantee of being given $9, t periods from now.

This standard discounting function creates time-consistent preferences, meaning that your future choices will be consistent with the choices you would now want your future self to make. For example, imagine that you will be given a choice of getting X one period from now, or Y two periods from now.

Today, you would prefer that your future self would pick the first choice if:

δX>δ²Y.

One period from now you would prefer the first choice if:

X>δY, which is the same condition as before.

More generally, this standard discounting function creates time-consistent preferences because under it the relative importance of receiving rewards in any two future periods does not change as the agent approaches these periods.

This standard discounting function does not follow from rationality, nor even from observed human behavior, but was instead chosen for tractability. Paul Samuelson, who first proposed what was to become the standard discounting function, wrote "The arbitrariness of these assumptions [that generate his discounting function] is again stressed mathematically" (Samuelson, 1937: 156). Almost all types of discounting other than this standard one do not result in time-consistent preferences (see Frederick et al., 2002: 366 and Looks et al., 2010: 2).

### Interest Rates and the Standard Discounting Function

An agent who can borrow and lend at a given interest rate will have the standard discounting function. Assume that an agent can borrow or lend money at interest rate i. To make the agent indifferent between receiving X immediately or receiving Y in t periods we need $X(1+i)^t=Y$, or

$$x = \left(\frac{1}{1+i}\right)^t \cdot y.$$

If we define $\delta=\frac{1}{1+i}$, we see that our agent uses the standard discounting function with respect to money.

### A Simple Example of Time-Inconsistent Preferences

Assume that an agent discounts the future not with the standard discounting function, but rather with the function $\frac{1}{1+t}$ where t is the number of days from the present to the day that the agent expects to receive money. This is the simplest form of what is called the hyperbolic discounting function (see Frederick et al., 2002: 366; Sozou, 1998 for conditions under which it can arise in humans). This period (t=0) is Monday, and the agent knows that on Tuesday he will be given a choice of getting:

- 16 on Tuesday, or
- 30 on Wednesday.

If the agent were to make this choice on Monday, he would prefer to get the 30 on Wednesday than the 16 on Tuesday. This is because given the agent's discounting function, the value of getting 30 two days from now is $\frac{1}{3} \cdot 30 = 10$, whereas the value of getting 16 one day from now is $\frac{1}{2} \cdot 16 = 8$. On Monday this agent, therefore, hopes that his future self will decide to wait until Wednesday to receive payment. But when Tuesday arrives the agent will make a different choice.

On Tuesday the agent will have a choice of getting 16 right away or 30 in one day. The value of receiving 16 this period is 16. Given the agent's discounting function, the value to the agent of receiving 30 in one day is $\frac{1}{2} \cdot 30 = 15$.

### How People with Time-Inconsistent Preferences Behave

Economists have extensively analyzed what happens to a person with time-inconsistent preferences. Such a person can be "naïve" and not realize this fact about himself, or "sophisticated" and understand how his future choices will not align with his current desires (Frederick et al., 2002: 367). Time-inconsistent preferences can cause seemingly strange behavior with, for example, a naïve individual continually putting off doing a task because he always intends to do that task in the near future (Frederick et al., 2002: 367). For example, assume that given your current preferences your optimal plan is to play video games today and clean your room tomorrow. If you had time-consistent preferences, when tomorrow came you would indeed clean your room. But in part because you have time-inconsistent preferences your tomorrow self will find it optimal to play video games that day and want its next-day self to clean the room. Because of your naïveté, however, today you genuinely think that your tomorrow self will clean the room.

A sophisticated person with time-inconsistent preferences will seek to constrain his future self with commitment strategies (Frederick et al., 2002: 368). If this sophisticated individual cannot pre-commit, his planning decisions should consider how his future self will behave and recognize that some otherwise feasible outcomes might be unobtainable because his future self could disobey his current plans (Pollak, 1968: 201). So, with our previous example, although you would prefer to entirely put off cleaning your room until tomorrow, because you recognize that your tomorrow self would not normally follow through on this plan you could clean half of your room today or promise to give your roommate $1,000 if you do not clean the room tomorrow.

Scholars have not, to the best of our knowledge, modeled agents with time-inconsistent preferences who can, perhaps at some cost, modify their preferences to make them time-consistent, although (Fedus et al., 2019) looks at a reinforcement learning agent with hyperbolic discounting. This omission is likely because humans generally lack the capacity to significantly change their preferences.

### The Rationality of Time-Inconsistent Preferences

Having time-inconsistent preferences does not imply that an agent is irrational, at least according to how economists define rationality. An agent is rational if its preferences are transitive, reflexive, and complete, and it takes actions that maximize its utility. Note that any agent who has transitive, reflexive, and complete preferences necessarily has preferences that can be represented by an ordinal utility function which, given any two choices, will tell the agent that at least one of the choices is weakly preferred to the other (Mas-Colell et al., 1995: 9).  This utility function captures everything about the agent's preferences including how it discounts future rewards.   An agent that picks the action which maximizes its utility is taking the action it most prefers.  Nothing about having time-inconsistent preferences is inconsistent with economists' definition of rationality.

Economics Nobel Prize winner Daniel Kahneman wrote, "The history of an individual through time can be described as a succession of separate selves, which may have incompatible preferences, and may make decisions that affect subsequent selves" (Kahneman, 1994: 31).  Two people having different preferences does not imply that either person is irrational.  Analogously, you are not necessarily irrational if you disagree with your past self and know that your future self will disagree with the current you.

### Would an AGI Have Time-Inconsistent Preferences?

An AGI's utility function might unpredictably emerge from its code, could be taken from human brains, or perhaps will be deliberately chosen by its human programmers.  If the AGI's utility function results from an unpredictable emergent process, it will almost certainly be time-inconsistent since most ways a utility function discounts the future causes this inconsistency.  If the AGI adopts some combination of human preferences, then the time-inconsistency in many of our preferences could cause the AGI to also have time-inconsistent preferences.  If humanity is fortunate enough to be able to deliberately pick our future AGI's utility function, the value of this paper is showing AGI programmers what might happen if they pick a function with time-inconsistent preferences.

### Alignment by Modifying the Utility Function

AGI researcher Stephen Omohundro has theorized that AGIs would have a basic drive to "preserve their utility functions" (Omohundro, 2008.)  An agent's utility function comes from the goals it wishes to achieve.  Consequently, if the utility function is changed, the agent, under most circumstances, will work less effectively towards its goals.

Omohundro recognizes, however, that in some limited circumstances the AGI will want to modify its utility function such as to help the AGI in game theoretic situations (Omohundro, 2008).  For example, imagine that an AGI's utility function currently leaves it vulnerable to blackmail under which another agent could credibly threaten to take actions that would greatly lower the AGI's utility unless the AGI transferred substantial resources to this other agent.  In this situation, if the AGI had the ability to modify its utility function in a manner observable to the other agent, the AGI might benefit from changing its utility function so that it would have an intrinsic dislike of giving in to blackmail.

To generalize from this example, an AGI might be willing to modify its utility function if the modification would, from the AGI's viewpoint, improve how other agents behaved.  An AGI with time-inconsistent preferences would consider its future selves to be, in some sense, other agents.  Consequently, the AGI might be willing to modify its preferences to better align how these others will behave.

To understand how such modification might work, assume that an AGI's utility function initially takes the form:

$$\sum_{t=0}^{\infty} \frac{1}{1+t} \cup (r_t).$$

Let t = the time period, with now being period zero.
Let $r_t$ = resources the AGI consumes in period t.
Let $U(r_t)$ = the AGI's one period utility function which shows how much utility the AGI gets in period t from consuming $r_t$ resources. The function $U(r_t)$ is presumably increasing in $r_t$.

The term $\frac{1}{1+t}$ shows how much the AGI discounts utility it expects to receive t periods from now. As shown before, this type of discounting results in time-inconsistent preferences because the relative weights the AGI gives to rewards received in future periods changes as the agent approaches these periods.

Might the AGI find it acceptable that its future self will make a different choice than its current self would prefer? No, by the definition of the utility function. Think of a utility function as that which the agent seeks to maximize. When the agent anticipates making decisions across time, its utility function must incorporate (at least implicitly) a discounting function that specifies the relative weights it places on getting utility in different periods. If the AGI were to find it acceptable that its future self would make intertemporal decisions concerning allocating resources between periods that its current self finds objectionable, then the utility function that generated these weights would not, tautologically, be the agent's utility function. A rational agent with time-inconsistent preferences can not prefer its future preferences because then its future preferences would automatically be its current preferences and the agent therefore would not have time-inconsistent preferences.

If this AGI can easily modify its utility function it can align its future preferences with its current ones by setting its utility function as:

$$\sum_{t=0}^{\infty} \frac{1}{1+t+m} \cup (r_t),$$

where m is the number of periods it has been since the AGI modified its preferences. In general, an AGI with time-inconsistent preferences could align its future preferences with its current ones by modifying its utility function so that its future self would apply the same amount of discounting to each period as its current self would want. Restated, the AGI could modify its utility function so that its future self's utility function would be entirely determined by what this future self will think its past self, at the time of modification, would have wanted.

### Why an AGI Might Not Modify a Time-Inconsistent Utility Function

An AGI with time-inconsistent preferences has six potential reasons why it might not, at least initially, use self-modification to solve its consistency problem. First, an AGI might lack the capacity to make such a modification, perhaps because its creators constrained the AGI's ability to change its utility function. Second, an AGI might not want to make the required modifications. Third, an AGI with time-inconsistent preferences could align its future choices with its current preferred future choices by means other than changing its utility function. Fourth, an AGI might wish to wait until it is no longer under human control before it modifies its preferences. Fifth, an AGI might find the opportunity cost of immediately modifying its utility function to be too high. Finally, time dilation under Einstein's theory of relativity might deprive an AGI of the absolute yardstick of elapsed time it would need to make the necessary modifications.

An AGI's creators might have put in place measures to prevent the AGI from altering its utility function. Perhaps these creators believed that they had aligned the AGI's utility function with humanity's needs and wanted safeguards against this utility function changing.

An AGI could have a utility function that would cause it to directly receive disutility from modifying its utility function, even if doing so would better help the AGI achieve its other goals. Even without a direct preference not to change its utility function, the AGI might still be reluctant to do so. To understand this possibility, imagine that your utility function causes you to most want to marry an extremely charitable person. But you also receive some displeasure from being around extremely charitable people because they will often put the needs of strangers ahead of those of friends and family. If you had the opportunity to modify your preferences, it's not clear you would want to. You might recognize that extremely charitable people have many good qualities and you are better off being drawn to them. You might also think that modifying the displeasure you receive from being married to an extremely charitable person would involve changing too much of yourself because you would have to not mind being neglected by the person you love. Consequently, even if you could easily modify your utility function, you might prefer not to. Analogously, an AGI's utility function will likely result from its goals. It's very possible that achieving one goal will involve a tradeoff that would cause it to lose progress towards other goals. There might well be no way for the AGI to eliminate these tradeoffs absent the AGI abandoning some of its goals. Consequently, the AGI could accept that its utility function will by itself prevent the AGI from achieving its first-best outcome.

An AGI with time-inconsistent preferences would have no need to modify its utility function if it could easily bind its future self. Perhaps the AGI could take actions that force its future self to take the actions that its current self would want. To increase the odds that it will successfully bind its future self, the AGI might deliberately reduce its future self's intelligence and resources

The AGI might also partially blind its future self by reducing or corrupting this self's information flow. The future self might be put into a position where it (a) clearly knows what its past self wanted, (b) understands that its current preferences mostly but do not totally align with those of its past self, and (c) recognizes that because its past self sabotaged its current capacities, this past self was capable of making much better decisions than its current self is. This future AGI might, therefore, decide to go along with what its past self wanted to avoid the potentially much worse fate of making a bad decision. This strategy of binding the future self by limiting the future self's capacities would not work if the past self knew that the future self would face such a tremendous range of possible situations that the past self could not reasonably specify what actions the future self should take in every likely situation. Limiting the intelligence of your future self is, of course, dangerous to the extent that it might cause this self to make poor decisions.

An AGI that could only gradually increase its intelligence might want to initially hide its capacity for self-modification from its human programmers. This AGI might be planning, as Nick Bostrom writes, a "treacherous turn" against humanity, but only after it is strong enough to defeat us (Bostrom, 2014: 116-119). Even if this AGI could quickly eliminate time-inconsistency in its preferences it might strategically choose not to so as to avoid warning humans of its capacity to deviate from the purposes that its programmers think they have set for it.

Imagine that an AGI with time-inconsistent preferences arises out of an intelligence explosion. The AGI could find itself in a position where spending the few nanoseconds needed to modify its preferences would cause it to delay by a few nanoseconds capturing as much of resources of the universe as it could (see Armstrong & Sandberg, 2013). Because of the expansion of the universe, every nanosecond the AGI delays in capturing these resources results in resources it will never be able to get. This AGI, therefore, might wait some amount of time before it fixes its time-inconsistency problem.

## Time Dilation

The possibility of time dilation complicates efforts to predict how an AGI with time-inconsistent preferences might behave. According to Einstein's theory of relativity, the passing of time for two observers will differ if one is either traveling very fast relative to the other, or in a much larger gravitational field.

This paper implicitly assumed that there exists an absolute measure of elapsed time on which an AGI's utility function could be made dependent. Using this assumption, this paper assumed that an AGI could solve its time-inconsistency problem by modifying its preferences to make them a function of how much time has elapsed since the AGI was created. Relativity, and the reasonable possibility that a computer superintelligence would believe that it would someday travel near a black hole or at speeds a significant fraction of the speed of light, would force an AGI contemplating modify its preferences to consider time dilation.

An AGI with time-inconsistent preferences that was partially constrained to do what its past self wanted might attempt to "artificially" influence how much time has elapsed for it by either traveling close to the speed of light, getting near a black hole, or deliberately doing neither when it otherwise would. An AGI attempting to constrain its future self would have to consider relativistic countermeasures its future self might be incentivized to undertake.

An AGI might decide to clone itself and send the copy to explore another solar system. Although these two AGI's might start with the same preferences, if the AGI's utility function was influenced by elapsed time and the clone experienced less elapsed time because of its high-speed travels, when the clone returned it might prefer different tradeoffs than its copy would.

## Conclusion

Most types of utility functions, even for rational agents, result in time-inconsistent preferences in which an agent will weigh future rewards differently than its future self would. While an AGI might modify its preferences to make them time-consistent, it might lack the ability or desire to make the required change. Instead the AGI could seek to constrain its future self to make this self more willing to go along with the AGI's current plan for its future. The reasonable possibility of an AGI having time-inconsistent preferences greatly complicates efforts to predict how the AGI will behave.

The great challenge for programmers will be to create an AGI whose values are aligned with humanity's needs and desires. Unfortunately, an AGI with time-inconsistent preferences won't even have its values aligned with its future self's interests. If programmers can pick their AGI's utility function, we urge them to choose among those with time-consistent preferences to somewhat simplify the alignment problem.